\documentclass[lettersize,journal]{IEEEtran}
\usepackage{amsmath,amsfonts}
\usepackage{algorithmic}
\usepackage[ruled, vlined, commentsnumbered, linesnumbered]{algorithm2e} % new package
\usepackage{array}
\usepackage[caption=false,font=normalsize,labelfont=sf,textfont=sf]{subfig}
\usepackage{textcomp}
\usepackage{stfloats}
\usepackage{url}
\usepackage{verbatim}
\usepackage{graphicx}
\usepackage{cite}
\usepackage{booktabs}
\usepackage{multirow}
\usepackage{xcolor}
\newcommand{\etal}{{\em et al.}}       % et al.
\newcommand{\eg}{{\em e.g.}}           % e.g.
\newcommand{\ie}{{\em i.e.}}           % i.e.
\newcommand{\Exp}{\widetilde{\operatorname{Exp}}}
\newcommand{\Log}{\widetilde{\operatorname{Log}}}

\newcommand{\sw}{\color{black}}
\newcommand{\ws}{\color{black}}

\begin{document}

\title{Deep Geometric Learning with Monotonicity Constraints for Alzheimer's Disease Progression}

\author{Seungwoo Jeong, Wonsik Jung, Junghyo Sohn, and Heung-Il Suk, \IEEEmembership{Senior Member, IEEE}
\thanks{Seungwoo Jeong and Junghyo Sohn are with the Department of Artificial Intelligence, Korea University, Seoul 02841, Republic of Korea (e-mail: sw\_jeong@korea.ac.kr; jhsohn0633@korea.ac.kr).}
\thanks{Wonsik Jung is with the Department of Brain and Cognitive Engineering, Korea University, Seoul 02841, Republic of Korea (e-mail: ssikjeong1@korea.ac.kr).}
\thanks{Heung-Il Suk is with the Department of Artificial Intelligence and the Department of Brain and Cognitive Engineering, Korea University, Seoul 02841, Republic of Korea and the corresponding author (e-mail: hisuk@korea.ac.kr)}
\thanks{Seungwoo and Wonsik have contributed equally to this work.}
}

\markboth{IEEE TRANSACTIONS ON NEURAL NETWORKS AND LEARNING SYSTEMS, VOL. XX, NO. XX, XXXX 2023}%
{Jeong \MakeLowercase{\textit{et al.}}: Deep Geometric Learning with Monotonicity Constraints for Alzheimer's Disease Progression}


\maketitle

\begin{abstract}
Alzheimer's disease (AD) is a devastating neurodegenerative condition that precedes progressive and irreversible dementia; thus, predicting its progression over time is vital for clinical diagnosis and treatment. Numerous studies have implemented structural magnetic resonance imaging (MRI) to model AD progression, focusing on three integral aspects: (i) temporal variability, (ii) incomplete observations, and (iii) temporal geometric characteristics. However, deep learning-based approaches regarding data variability and sparsity have yet to consider inherent geometrical properties sufficiently. The ordinary differential equation-based geometric modeling method (ODE-RGRU) has recently emerged as a promising strategy for modeling time-series data by intertwining a recurrent neural network and an ODE in Riemannian space. 
Despite its achievements, ODE-RGRU encounters limitations when extrapolating positive definite symmetric matrices from incomplete samples, leading to feature reverse occurrences that are particularly problematic, especially within the clinical facet.
Therefore, this study proposes a novel geometric learning approach that models longitudinal MRI biomarkers and cognitive scores by combining three modules: topological space shift, ODE-RGRU, and trajectory estimation. We have also developed a training algorithm that integrates manifold mapping with monotonicity constraints to reflect measurement transition irreversibility. We verify our proposed method’s efficacy by predicting clinical labels and cognitive scores over time in regular and irregular settings. Furthermore, we thoroughly analyze our proposed framework through an ablation study. 
\end{abstract}

\begin{IEEEkeywords}
Alzheimer's disease, longitudinal data, missing value imputation, neural ordinary differential equations, geometric modeling
\end{IEEEkeywords}

\section{Introduction}
\label{sec:introduction}
\IEEEPARstart{A}{lzheimer}'s disease (AD) is a degenerative neurological condition hallmarked by an irreversible and gradual cognitive descent into dementia, featuring memory loss, impaired movement, mild cognitive impairment (MCI), and other related symptoms~\cite{apostolova2008mapping}. Identifying potential biomarkers during presymptomatic stages is crucial for effective treatment; therefore, predicting accurate clinical status and changes over time is paramount.

Deep learning utilizing magnetic resonance imaging (MRI) is the leading technique for modeling AD progression. Among these algorithms, recurrent neural network (RNN)-based approaches notably encapsulate temporal brain morphology or pathology changes~\cite{hochreiter1997long, cho2014learning}. Although deep learning models have attested to remarkable performance in predicting AD progression using regular and complete observed samples, they are often hindered by sparse or irregular data in genuine clinical settings. Previous studies aiming to resolve this limitation have proposed several imputation techniques to generate complete data by filling in the missing values~\cite{yoon2018estimating,nguyen2020predicting,jung2021deep}. A novel and intriguing strategy for addressing irregularly sampled time-series data is through ordinary differential equations (ODEs)~\cite{chen2018neural, rubanova2019latent, de2019gru, nazarovs2022mixed, jeong2021efficient}. ODEs model hidden state dynamics over time rather than the input data directly, which is particularly effective for irregularly sampled time-series data, where time intervals between observations are inconsistent. By formulating the problem through ODEs, the model learns the data’s underlying continuous-time dynamics even with missing or irregularly spaced observations.

% In recent years, various techniques have endeavored to simultaneously capture geometric and continuous time-series data characteristics~\cite{de2019gru, nazarovs2022mixed, jeong2021efficient}. Specifically, ODE-RGRU~\cite{jeong2021efficient} interweaves RNNs and an ODE on the symmetric positive-definite (SPD) space to enhance time-series data performances, such as sensor data, electroencephalograms, and videos. However, ODE-RGRU requires complete observations for mapping onto an SPD space with covariance estimation; thus, sparse or incomplete data limits its effectiveness.

\begin{figure}[t]
\centering
    \includegraphics[width=.55\textwidth]{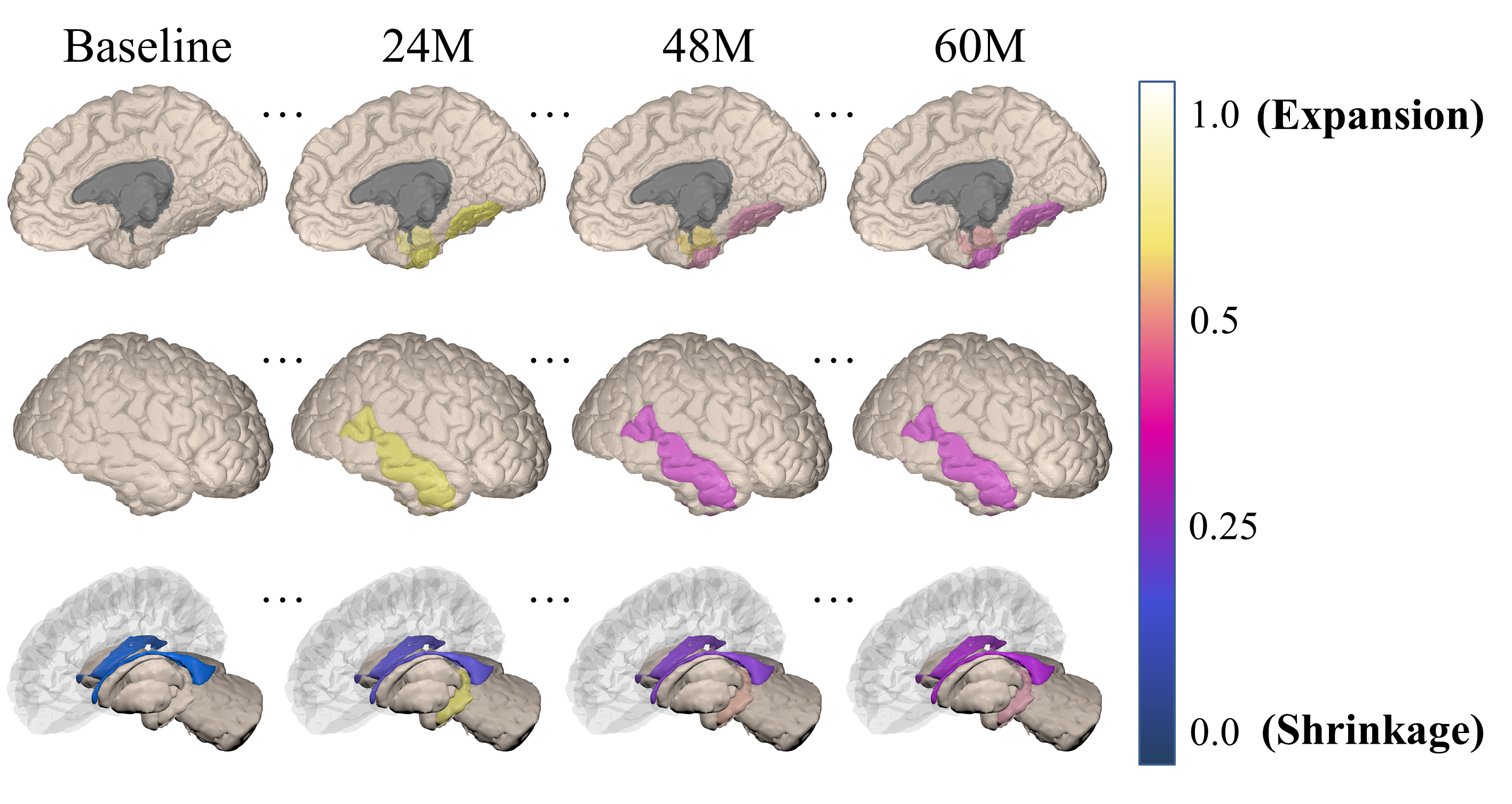}
    \caption{\ws Depiction of temporal morphological changes in brain regions from a clinical perspective within participants. The color-coded values denote the extent of volume changes compared to a first-visit sequence (\eg, baseline). The dark navy or beige represents the shrinkage of volume or enlargement relative to the first-visit sequence.}
    \label{fig:monotonic}
    \label{fig:introduction}
\end{figure}

{\sw In recent years, there has been a growing interest in developing techniques that can effectively capture the geometric and continuous characteristics of time-series data~\cite{de2019gru, nazarovs2022mixed, jeong2021efficient}. One such approach, known as ODE-RGRU~\cite{jeong2021efficient}, combines RNNs and an ODE on the symmetric positive-definite (SPD) space to improve the performance of time-series data analysis. This method has shown promise in applications involving sensor data, electroencephalograms, and videos. However, a significant limitation of ODE-RGRU is its reliance on complete observations to map data onto the SPD space with covariance estimation. As a result, its effectiveness suffers when dealing with sparse or incomplete datasets. This presents a problem in real-world circumstances where data is not always readily available.
Furthermore, ODE-RGRU cannot accurately depict clinical attributes related to monotonically growing or decreasing aspects, which are especially essential in the context of AD progression depicted in Fig.~\ref{fig:monotonic}.}

This work proposes a novel architecture for modeling AD progression while leveraging sporadic observations assembled upon the ODE-RGRU~\cite{jeong2021efficient}. We aim to surmount three critical challenges in modeling longitudinal MRI and cognitive scores: (i) capturing temporal feature variability, (ii) managing sporadic observations, and (iii) ensuring geometric temporal data continuity. Our proposed framework comprises a topological space shift module, ODE-RGRU, and trajectory estimation module, each serving a distinct purpose: the topological space shift transforms data into Cholesky space to enable geometric representation, ODE-RGRU learns hidden state trajectories to allow for continuous modeling, and the trajectory estimation module ascertains missing values in incomplete samples. 
{\sw Additionally, we introduce a training algorithm that integrates monotonicity constraints into the manifolds mapped from missing observations. This integration prevents clinically improbable inverse transitions between features and enables estimation by utilizing hidden state trajectories obtained from completed observations.}
Our proposed framework is evaluated through longitudinal cohorts centered on clinical status classification and cognitive score prediction. We also completed diverse analyses to verify the framework’s effectiveness, demonstrating its ability to capture intricate longitudinal MRI biomarkers and cognitive score data dynamics. 

The main contributions of this work are as follows:
\begin{itemize}
    \item We devise a novel geometric learning framework that leverages temporal variability, incomplete observations, and geometrical longitudinal data properties to model AD progression.
    \item We develop a training algorithm to meld monotonicity constraints with a manifold mapped from missing observations, thereby preventing a reverse transition case and enabling estimation by applying hidden state trajectories from completed observations.
    \item We verify our proposed framework’s efficacy by accomplishing extensive analyses using publicly available longitudinal data from The Alzheimer’s Disease Prediction of Longitudinal Evolution (TADPOLE).
\end{itemize}

% This work augments our research presented at the 22nd IEEE International Conference on Data Mining~\cite{jeong2022deep}. 
{\sw This work is an extension of the previous conference version~\cite{jeong2022deep}.}
We supplemented our original work by incorporating monotonicity constraints into the training process to reflect clinical MRI biomarker irreversibility. Additionally, we conducted an ablation study and an irregular-time setting analysis to certify our proposed framework’s versatility and applicability.

\section{Related Work}
\label{sec:relatedwork}
\subsection{RNN-based AD Progression Modeling}
Deep learning with RNN-based methodologies has demonstrated remarkable potential in modeling disease progression, such as LSTM~\cite{hochreiter1997long} and GRU~\cite{cho2014learning}, widely used to capture temporal patterns from time series data. These models efficiently curb the issue of vanishing and exploding gradients and capture long-term dependencies, befitting disease progression modeling.

RNNs effectively model discrete-time dynamical systems with regular input and output time intervals. However, they are trammeled by incomplete data bearing unpredictable acquisition timings, effectuating sparse and missing data. Incomplete data is a notorious vexation for standard RNN-based techniques. In response, previous studies either remove missing observations~\cite{ghazi2019training} and utilize masks concerning missing observations or apply missing value imputation techniques~\cite{yoon2018estimating,nguyen2020predicting,jung2021deep} to beget complete data. Ghazi \etal~\cite{ghazi2019training} introduced a Peephole LSTM model that requires zero imputation for marking missing observations when estimating MRI biomarkers and a Linear Discriminant Analysis (LDA) classifier for systemization. Meanwhile, Nguyen \etal~\cite{nguyen2020predicting} employed a MinimalRNN to predict disease progression and directed a trained model to impute missing dataset values. The authors utilized support vector regression on continuous variables and support vector machines to predict categorical variables simultaneously. Yoon \etal~\cite{yoon2018estimating} handled electronic health record data and bi-directional RNNs to impute missing variables by considering temporal relations. Similarly, \etal~\cite{jung2021deep} proposed an imputation technique that leverages imputed variables' temporal and spatial relations from available observations. While these methods have promise in managing incomplete data, certain RNN characteristics, such as fixed time intervals, remain a concern.

Recent methods have modeled hidden state dynamic patterns to surpass the constraints of irregular time-series data. For example, neural ODE~\cite{chen2018neural} parameterizes hidden state derivatives by regarding time as a variable, subsequently solving the initial value problem. Moreover, Latent ODE~\cite{rubanova2019latent} and its variants~\cite{de2019gru, kidger2020neural, yildiz2019ode2vae, chen2021continuous} are recommended for handling irregularly-sampled time-series data. In particular, ME-NODE~\cite{nazarovs2022mixed} has proven its capability in analyzing AD progression through a probabilistic model incorporating mixed effects.

\subsection{Geometric Modeling}
Recent research has established that geometric modeling can effectively derive geometric data characteristics that pertain to the SPD matrix. For instance, \cite{huang2017riemannian} the authors designed and named the Riemannian network SPDNet for non-linear SPD matrix learning. This network includes transformation, non-linear activation, and output layers that satisfy the SPD matrix assumption. \cite{brooks2019riemannian} One study pioneered Riemannian batch normalization, which enhances training stability and performance by applying the Riemannian Fréchet mean, parallel transport, and non-linear structured matrix transformation. 

In addition, several proposed approaches augment geometric modeling to handle multivariate time-series data. For instance, SPDSRU~\cite{chakraborty2018statistical} fabricates a statistical recurrent network that harnesses non-Euclidean temporal, longitudinal, and ordered data. Alternatively, ManifoldDCNN~\cite{zhen2019dilated} redefines dilated convolutional networks in the Riemannian manifold to contend with technical and computational challenges. Gruffaz \etal~\cite{gruffaz2021learning} launched a mixed-effect Riemannian metric learning method that models disease progression by disentangling time and space variability. Furthermore, ODE-RGRU~\cite{jeong2021efficient} unifies the RNN in Cholesky space and the manifold ODE for continuous modeling in manifold space. Although ODE-RGRU exhibits impressive accomplishments with various time-series data, it is constrained by the SPD representation and encounters issues when faced with missing observations at specific timesteps.

\begin{figure*}[t]
\centering
    \includegraphics[width=1\textwidth]{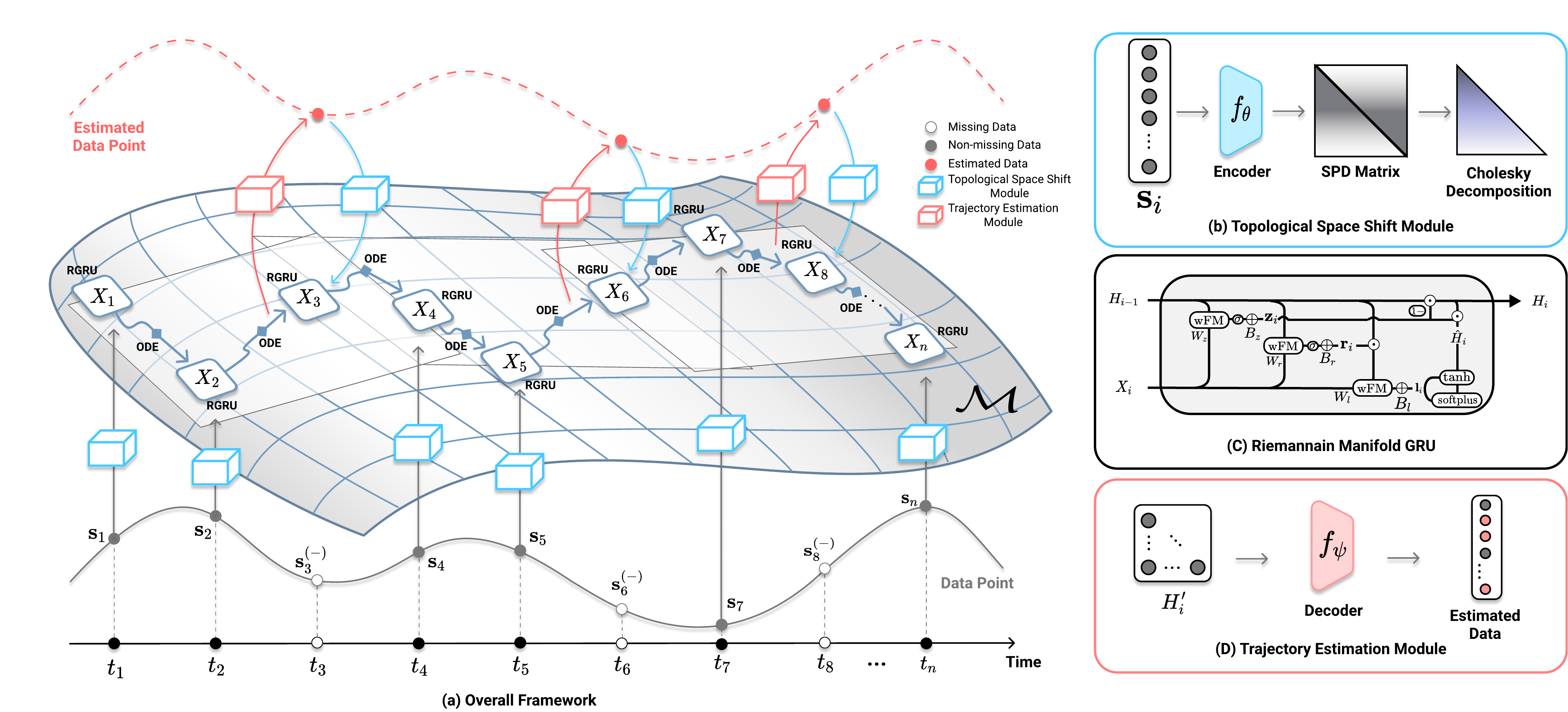}
    \caption{
    The proposed framework incorporates three primary modules: topological space shift, ODE-RGRU, and trajectory estimation. First, the topological space shift module transforms $S_T$ into a manifold point $\mathcal{X}_T$. Next, the RGRU calculates the hidden state, which maps the estimated points into the ODE solver's tangent space using $\Log(\cdot)$. The hidden state is then projected into the Cholesky space using $\Exp(\cdot)$. 
    Finally, the decoder $f_\psi$ takes the previous time point's hidden state as input, which contains the observed data up to that point, and estimates the missing values. The estimated values are then used as inputs to the topological space shift module to generate the manifold point $\mathcal{X}_t$.
    }
    \label{fig:fig_framework}
\end{figure*}

\section{Proposed Method}
\label{sec:proposed}
\vspace{5pt}\noindent\textbf{Notations.}
The input sequence was represented as $S_T \equiv \{\mathbf{s}_{1}, \mathbf{s}_{2}, \mathbf{s}^{(-)}_{3}, \cdots, \mathbf{s}_{T}\}$, where each $\mathbf{s}_i \in \mathbb{R}^{c}$ is a one-dimensional vector containing $c$ variables. The notation $\mathbf{s}^{(-)}_i$ denotes an incomplete vector with observations missing for certain variables. 
A set of SPD matrices are used in the manifold space at each time step $t\in\{1,2,\dots, T\}$ by $\mathcal{C}_T \equiv \{C_{1}, C_{2}, \cdots, C_{T}\}$ and the corresponding diffeomorphism matrix in the Cholesky space by $\mathcal{X}_T \equiv \{X_{1}, X_{2},\cdots, X_{T}\}$.

\vspace{5pt}\noindent\textbf{Problem Statement.} 
Given the input sequence $S_T$, modeling AD progression aims to predict disease status shifts over time, including cognitively normal (CN), MCI, and AD. Each input data timepoint includes MR volumetric information and cognitive test scores potentially lacking observations. Our framework exercises autoregressive modeling with monotonicity constraints to estimate missing values, effectively preventing reverse state transitions.

\subsection{ODE-RGRU} 
ODE-RGRU~\cite{jeong2021efficient} is a novel SPD matrix learning framework that combines manifold ordinary differential equations and RNNs. The authors aim to surmount the rigid constraints in learning SPD matrices through a diffeomorphism mapping technique. This method maps Riemannian manifolds onto a Cholesky space, allowing for more flexible and efficient SPD matrix parameterization.

\vspace{5pt}\noindent\textbf{Cholesky Space.} 
The Cholesky space ($\mathcal{L}_+, g$) is a smooth, real submanifold that the Cholesky Riemannian manifold decomposition can represent. The notation $\lfloor \cdot \rfloor$ strictly signifies the matrix’s lower triangular part, and $\mathcal{D}(\cdot)$ refers to the diagonal. Represented by $g$, the Riemannian metric is calculated by adding the products of the two matrices' elements:
\begin{align}
    g_L(X, Y) = \sum_{i>j}X_{ij}Y_{ij} + \sum_{j}X_{jj}Y_{jj}L^{-2}_{jj}
    \label{eq1}
\end{align}
where $X, Y \in \mathcal{L}$ and $L \in \mathcal{L}_+$.

\vspace{5pt}\noindent\textbf{Proposition 1. }\cite{lin2019riemannian} \textit{The Cholesky map is a diffeomorphism between $\mathcal{L}_+$ and $\mathcal{S}^+_d$ smooth manifolds.}

\noindent Proposition 1 establishes that the Cholesky map is a one-to-one and onto function with a differentiable inverse, implying that training with a deep neural network (NN) is plausible.

\vspace{5pt}\noindent\textbf{Exponential and Logarithmic Operations. } 
Exponential and logarithmic operations control mapping between manifold and tangent spaces; when defined in the Cholesky space~\cite{lin2019riemannian}, they are computationally more efficient than conventional Riemannian operations.
\begin{align}
    \label{eq2}
    \Exp_L(X) &= \lfloor L\rfloor + \lfloor X\rfloor + \mathcal{D}(L)\exp\{\mathcal{D}(X)\mathcal{D}(L)^{-1}\}, \\
    \label{eq3}
    \Log_X(L) &= \lfloor L\rfloor - \lfloor X \rfloor + \mathcal{D}(X)\log\{\mathcal{D}(X)^{-1}\mathcal{D}(L)\}
\end{align}

\vspace{5pt}\noindent\textbf{Fr\'echet Mean. } 
The Fr\'echet mean generalizes the Euclidean mean or expected probability distribution commonly employed in numerous operations (\eg, attention~\cite{vaswani2017attention}). However, the standard Fr\'echet mean formulation entails an \textit{arg}min operation that is not amenable to a closed-form solution. It is typically computed with an iterative solver, such as the Karcher flow~\cite{karcher1977riemannian}, which operates in the tangent space before returning to the manifold. However, in the Cholesky space, the Fr\'echet mean, also called the \textit{log-Cholesky} mean $\mu_{\mathcal{L}_+}$~\cite{lin2019riemannian}, has a closed-form solution that makes it computationally efficient to compute:
\begin{align}
    \label{eq4}
    \mu_{\mathcal{L}_+} = \frac{1}{N}\sum^N_{i=1}\lfloor X_i \rfloor + \exp\Big\{N^{-1}\sum^N_{i=1}\log \mathcal{D}(X_i) \Big\}.
\end{align}

\vspace{5pt}\noindent\textbf{Manifold ODE. }  
Manifold ODEs can solve the initial value problem by integrating a vector field $f$ over a curve $\mathbf{v}: [t_s, t_e] \rightarrow \mathcal{M}$, where $\mathcal{M}$ is the manifold, as described in~\cite{hairer2011solving}:
\begin{align}
    \label{eq5}
    \frac{d\mathbf{v}(t)}{dt} = f(\mathbf{v}(t), t) \in  \mathcal{T}_{\mathbf{z}(t)}\mathcal{M}.
\end{align}
On manifold $\mathcal{M}$, the differential equation's solution $\mathbf{z}$ is given by $\mathbf{z}(t_e)$ at the end of the curve, with $\mathbf{z}(t_s)$ as the initial condition. The vector field $f$ is defined on $\mathcal{M}$, and the derivative $\frac{d\mathbf{v}(t)}{dt}$ lies within the tangent space $\mathcal{T}_{\mathbf{z}(t)}\mathcal{M}$ for all $t\in[t_s, t_e]$. 

Our proposed framework includes three modules to address the critical challenges in establishing longitudinal MRI and cognitive scores for AD. These modules utilize the previously described operations and manifold space definitions to achieve their respective goals.

\subsection{Topological Space Shift}
The first module encompasses the relationship between variables in longitudinal data. First, the data is transformed into a second-order feature statistic (i.e., covariance). Then, the convolutional neural network $f_\theta$ employs a shrinkage estimator~\cite{chen2010shrinkage} to ascertain second-order feature statistics as an SPD matrix $C$ on the manifold space. Next, we performed Cholesky decomposition on the SPD matrix $C = XX^\top$, where $X$ satisfies the positive diagonal component constraints. However, missing observations will impede direct SPD matrix estimations; therefore, estimation techniques are utilized to fill in the missing values. Even so, conventional methods that rely on simple statistical associations or linear functions do not consider geometrical characteristics.

\subsection{RGRU}
Unlike conventional RNNs that operate on Euclidean space, the RNN-based model variant with a gating mechanism (or the Riemannian manifold GRU [RGRU]~\cite{jeong2021efficient}) operates on Riemannian manifold space for more flexibility when modeling temporal complex data dynamics~\cite{zhen2019dilated, chakraborty2018statistical, chakraborty2020manifoldnet}. RGRU embodies three components:
\begin{enumerate}
    \item Weighted Fr\'echet mean (wFM): a \eqref{eq4} generalization with arbitrary re-weighting, defined as follows: 
    \begin{align}
        \label{eq6}
        \operatorname{wFM}&(\left\{X_i\right\}_{i=1,\cdots, N}, \mathbf{w}\in\mathbb{R}^{N}_{\geq0})  \nonumber\\
        = \frac{1}{N}&\sum^N_{i=1}(w_i \cdot \lfloor X_i \rfloor)
        + \exp\Big\{N^{-1}\sum^N_{i=1}w_i\cdot\log \mathcal{D}(X_i) \Big\}.
    \end{align}
    Here, $\mathbf{w}$ represents the weight consisting of non-negative real values.
    \item Bias addition ($\oplus$):
    \begin{align}
        \label{eq7}
        X \oplus Y = \lfloor X\rfloor + \lfloor Y\rfloor + \mathcal{D}(X)\mathcal{D}(Y)
    \end{align}    
    \item Non-linearity: this function must fulfill the Cholesky space constraint through a single activation function, such as a sigmoid, or two independent activation functions~\cite{jeong2021efficient}.
\end{enumerate}
The RGRU is formulated as follows: 
\begin{align}
    \label{eq8}
    \left\{\begin{array}{c l}
    \mathbf{z}_i &= \sigma(\text{wFM}(\left\{ X_i, H_{i-1}\right\}, W_{z}) \oplus B_z), \\
    \mathbf{r}_i &= \sigma(\text{wFM}(\left\{X_i, H_{i-1}\right\}, W_{r}) \oplus B_r), \\
    \mathbf{l}_i &= \text{wFM}(\left\{X_i, \mathbf{r}_i\odot H_{i-1}\right\}, W_{l}) \oplus B_l, \\
    H_{i} &= (1-\mathbf{z}_i) \odot H_{i-1} + \mathbf{z}_i \odot \hat{H}_i,
    \end{array}\right.
\end{align}
$H_i$ and $\hat{H}_i$ denote the current and candidate hidden states, respectively.  
$\mathbf{z}_i$ and $\mathbf{r}_i,$ represent the update and reset gate, respectively.
$W_{\{z,r,l\}}$ and $B_{\{z,r,l\}}$ are weight and bias parameters of RGRU. $\sigma$, $\operatorname{tanh}(\cdot)$, and $\operatorname{softplus}(\cdot)$ are sigmoid, hyperbolic tangent, and softplus activation functions, respectively. $\odot$ is the element-wise multiplication.
For simplicity, RGRU can be expressed as follows:
\begin{align}
    \label{eq9}
    H_i = \operatorname{RGRU}(X_i, H_{i-1}).
\end{align}

\subsection{Neural Manifold ODE}
Neural manifold ODEs~\cite{lou2020neural} were included for continuous modeling and estimating missing values' hidden states. The manifold ODEs consist of forward and backward processes defined in distinct spaces. An implicit method~\cite {hairer2011solving, chakraborty2018statistical} based on a step-based method~\cite{crouch1993numerical} executed the forward pass. Specifically, this involved the Riemannian exponential map~\cite{bielecki2002estimation} with the Euler method solver:
\begin{align}
    \label{eq10}
    H_{t+\epsilon} = \operatorname{exp}_{H_t}(\epsilon f_\phi(H_t, t))
\end{align}
$f_\phi$ represents an NN, and $\epsilon$ denotes a discrete time point.

In differential geometry, the mapping function derivative between two manifolds is defined as a function between the tangent space $D_xf: \mathcal{T}_x\mathcal{M} \rightarrow \mathcal{T}_x\mathcal{N}$. A previous study~\cite{lou2020neural} introduced a manifold-based adjoint sensitivity method~\cite{chen2018neural} constructed using \eqref{eq2} and \eqref{eq3}.
\begin{align}
    \label{eq11}
    \frac{d\Log(H_t)}{dt} = D_{\Exp(H_t)}\Log\Big(f_{\phi}(\Exp(H_t), t)\Big)
\end{align}
By updating $H_t$ to $H_{t+\epsilon}$, we utilize the updated state to estimate the missing data point.

\subsection{Trajectory Estimation}
ODE-RGRU was employed for time-series modeling, effectively capturing the data's geometric structure in the manifold space. 
% However, managing the topological space shift with sparse data and missing values is challenging. Missing data prompts information loss, which hinders capturing the data’s underlying manifold structure. Therefore, we propose a trajectory estimation module to estimate missing values and leverage them as inputs for the topological space shift module. 
{\sw Nevertheless, dealing with topological space shift modules when confronted with sparse data and missing values poses significant challenges. The absence of data leads to a loss of crucial information, hindering the ability to capture the underlying manifold structure of the dataset. To address this issue, we put forward a trajectory estimation module to estimate the missing values and employ them as input to the topological space shift module. By incorporating this approach into the model's learning process, we can effectively estimate the missing data through trajectory $H$, enhancing the model's representation power. By taking advantage of trajectory estimation, our proposed method handles missing data more effectively and contributes to overall improved performance in the learning process.}

Given data at time $t_i$ and $t_{i+1}$ with missing observations, denoted as $\mathbf{s}_{i}$ and $\mathbf{s}^{(-)}_{i+1}$, the trajectory estimation first calculates the hidden state $H_i$ using $\mathbf{s}_i$. Next, the ODESolver estimates the hidden state at time $t_i$ based on the available information and then estimates the hidden state $H_{i+1}$ by leveraging the learned trajectory. Finally, the decoder $f_\psi$ predicts the missing time point value. Subsequently, the decoder $f_\psi$ improves ODESolver’s representation capacity by estimating missing values and passing them back. {\sw We use a fully connected layer for the decoder $f_\psi$.}

\subsection{Training Algorithm}
Fig.~\ref{fig:fig_framework} illustrates our proposed framework’s architectural details. During the training process, a time-dependent dataset $\{(\mathbf{s}_1, t_1), (\mathbf{s}_2, t_2), \cdots\}$ is fed into the topological space shift module. The input $\mathbf{s}_i$ is transformed into a matrix $X_i$, corresponding to a point in the Cholesky space. $X_i$ is then fed into the ODE-RGRU, which combines a manifold ODE and an RGRU. Specifically, the ODE-RGRU solves an ODE with an RGRU unit by applying the following equations:
\begin{align}
    \label{eq12}
    H^\prime_i &= \text{ODESolve}(f_\phi, \Log(H_{i-1}), (t_{i-1}, t_i))
    \\
    \label{eq13}
    H_i &= \text{RGRU}(X_i, \Exp(H^\prime_i)),
\end{align}

Assuming that the missing value is at time point $t_{i+1}$, it is difficult to feed into the topological space shift. Therefore, we estimate $\mathbf{s}_{i+1}$ with $H^\prime_{i+1}$ using the trajectory estimation module. $H^\prime_{i+1}$ is obtained from \eqref{eq12}. The estimated $\mathbf{s}_{i+1}$ imputes missing values at the time point $t_{i+1}$ while preserving the observed points in $\mathbf{s}^{(-)}_{i+1}$.
\begin{align}
    \label{eq14}
    \mathbf{\hat{s}}_{i+1} = \mathbf{m}_{i+1} \odot \mathbf{s}^{(-)}_{i+1} + (1-\mathbf{m}_{i+1}) \odot f_\psi(H^\prime_{i+1})
\end{align}
where $\mathbf{m}_i$ is an indicator vector that specifies the missing observations. 
By learning the estimated data representation and the following ODE-determined trajectory, the decoder $f_\psi$ imputes these missing observations. The hidden state $H_{i+1}$ is calculated based on the estimated $\mathbf{\hat{s}}_{i+1}$ using the same process. Lastly, a linear layer and logistic regression classify the current prediction.
\begin{align}
    \hat{y}_{i+1} = \text{softmax}(W_y\cdot\Log(H_{i+1}) + \mathbf{b}_y).
\end{align}
Here, $\hat{y}_{i+1}$ represents the predicted class label and $W_y$ and $\mathbf{b}_y$ are learnable parameters.
Algorithm~\ref{alg:algorithm} summarizes the overall procedure of our proposed method.

\begin{algorithm}[t]
    \caption{Pseudo algorithm for the proposed framework}
    \label{alg:algorithm}
    \SetKwInOut{Input}{Input}
    \SetKwInOut{Output}{Output}
    \Input{Training dataset $\{(\mathbf{s}_i, t_i, m_i)\}_{i=1\ldots T}$; initialized network parameters $\theta,\;\phi,\;\psi$}
    \Output{Prediction $\{y_i\}_{i=1\ldots T}$}
    \begin{algorithmic}[1]
    \STATE Initial hidden state $H_0 = \mathbf{I}$ \tcp{Identity matrix}
    \FOR {$i$ in $1, 2, \cdots, T$}
        \IF{$\mathbf{s}_i$}
            \STATE $C_i = \operatorname{Shrinkage\;estimation}(f_\theta(\mathbf{s}_i))$
            \STATE $X_i X^\top_i = \operatorname{Cholesky\;decomposition} (C_i)$
            \STATE $H^\prime_i = \operatorname{ODESolve}(f_\phi, \Log(H_{i-1}), (t_{i-1}, t_i))$ %\tcp{Eqs.(\ref{eq3},\ref{eq12})}
            \STATE $H_i = \operatorname{RGRU}(X_i, \Exp(H_{i-1}))$ %\tcp{Eqs.(\ref{eq2},\ref{eq13})}
        \ELSIF{$\mathbf{s}^{(-)}_i$}
            \STATE $\mathbf{\hat{s}}_{i} = \mathbf{m}_{i} \odot \mathbf{s}^{(-)}_{i} + (1-\mathbf{m}_{i}) \odot f_\psi(H^\prime_{i})$ %\tcp{Eq.(\ref{eq14})}
            \STATE Return to step 4 and forward the process using estimated data point $\mathbf{\hat{s}}_i$
    \ENDIF
    \ENDFOR
    \STATE $\{y_i\}_{i=1\ldots T} =$ OutputNN($\Log(H_i)$) for all $i=1\ldots T$
    \STATE \textbf{return} $\{y_i\}_{i=1\ldots T}$ 
    \end{algorithmic}
\end{algorithm}

The proposed method’s training is facilitated by two loss functions with monotonic regularization for predicting cognitive scores and clinical outcomes. In addition, the simultaneous loss function optimization enhances the proposed method’s ability for underlying data representation.

\vspace{5pt}\noindent\textbf{Estimation Loss.} 
$\mathcal{L}_{\text{estim}}$ computes the correspondence between model predictions $\mathbf{\hat{s}}_{i+1}$ and ground-truth measurements $\mathbf{s}_{i+1}$ using the indicator vector $\mathbf{m}_i$. This operation ascertains estimated data points via the decoder $f_\psi$ and enhances ODE’s representation capability:
\begin{align}
    \mathcal{L}_{\text{estim}} = \sum^{T-1}_{i=1}\big(\mathbf{s}_{i+1} \odot \mathbf{m}_{i+1} - \mathbf{\hat{s}}_{i+1} \odot \mathbf{m}_{i+1} \big)^2.
\end{align}

\vspace{5pt}\noindent\textbf{Prediction Loss.} 
Our proposed method addresses the data imbalance by implementing the focal cross-entropy loss $\mathcal{L}_{\text{pred}}$~\cite{lin2017focal}:
\begin{align}
    \mathcal{L}_{\text{pred}} = -\sum^{T-1}_{i=1}\big[ \sum^K_{k=1} y_i(k)(1-\hat{y}_i(k))^\delta \log(\hat{y}_i(k)) \big]
    \label{eq:17}
\end{align}
where $\delta$ is a hyperparameter $(\delta \geq 0)$.

% Clinically, MRI biomarkers for training mirror irreversibility characteristics of the clinical status; thus, monotonicity must be considered when imputing incomplete data. Optimization forces a trend between previous and following values, serving as a monotonic function regularizer and preventing reverse transition.
{\sw Clinically, MRI biomarkers for training reflect the irreversibility of clinical status; consequently, monotonicity must be considered when imputing partial data. Our optimization strategy forces a trend between prior and subsequent values, acting as a monotonic function regularizer and preventing reversal.}
Therefore, we define the overall loss function $\mathcal{L}_{\text{total}}$ as follows:
\begin{align}
    \mathcal{L}_{\text{total}} = \lambda_1 \mathcal{L}_{\text{estim}} + \lambda_2 \mathcal{L}_{\text{pred}} + \lambda_3 ||\sum_i\operatorname{sgn}(\mathbf{s}_i - \mathbf{s}_{i-1})||
    \label{eq:18}
\end{align}
where $\lambda_1$, $\lambda_2$, and $\lambda_3$ are the hyperparameters to weight the corresponding losses, $\operatorname{sgn}(\cdot)$ denotes the sign function.

\section{Experiments}
\label{sec:exp}
This section details the dataset, preprocessing, experimental, and competing method settings for all experiments. Moreover, we discuss the experimental results from our proposed framework and comparative methods trialed on publicly available datasets. For detail, our implementation code exploited in experiments is available on GitHub\footnote{https://github.com/ku-milab/Deep-Geometric-AD}.

\subsection{Datasets and Preprocessing}
We collected the TADPOLE database\footnote{https://tadpole.grand-challenge.org/Data/.} regarding the Alzheimer's Disease Neuroimaging Initiative (ADNI) cohort, comprising data from 1,737 patients and 1,500 biomarkers compiled across 12,741 visits spanning 22 periods~\cite{marinescu2018tadpole,jung2019unified,jeong2022deep}.
Although TADPOLE provides numerous AD spectrum prediction biomarkers, this study opted to abide by previous studies~\cite{ghazi2019training,jung2019unified,jung2021deep} and selected six volumetric MRI features: entorhinal cortex, hippocampus, fusiform gyrus, middle temporal gyrus, ventricles, and whole brain. Moreover, we utilized both T1-weighted MRI scans and cognitive test scores including the mini-mental state exam (MMSE), Alzheimer's disease assessment scale (ADAS)-cog11, and ADAS-cog13, which were extracted from our collected dataset. Based on previous studies~\cite{ghazi2019training,jung2019unified,jung2021deep}, we divided subjects into three groups: CN and Significant Memory Concern (SMC), early MCI (EMCI) and late MCI (LMCI), and AD. 

We selected 11 of the 22 AD-progression prediction time sequences for a fair experimental comparison. Subjects without baseline or less than three visits were excluded, resulting in 691 subjects. While the competing methods could only be trained through traditional settings, our proposed framework can train in unconventional conditions. Therefore, our proposed framework used all 22 visits to compare performance with conventional settings. Due to the subjects’ brain volume and size differences, we normalized each MRI feature by the respective intra-cranial volume (ICV)~\cite{davis1977new}. In addition, we linearly normalized each MRI feature relative to its minimum and maximum values and normalized each cognitive score by dividing it by its maximum value, resulting in all values ranging between [0,1].

\begin{table}[!t]
    \centering
    \caption{Performance (mean$\pm$std) of a multi-class classification task in a longitudinal scenario. ($*$: $p<0.05$)}
    \label{exp:t1}
    \scalebox{.95}{
    \begin{tabular}{cccc}
    \toprule
    Method & mAUC & Recall & Precision \\
    \midrule\midrule
    LSTM-M & 0.758$\pm$0.054$^*$ & 0.596$\pm$0.090 & 0.537$\pm$0.162 \\ 
    LSTM-F & 0.740$\pm$0.039$^*$ & 0.535$\pm$0.092 & 0.562$\pm$0.127 \\ 
    MRNN \cite{yoon2018estimating} & 0.774$\pm$0.045$^*$ & 0.611$\pm$0.045 & 0.580$\pm$0.092 \\ 
    PLSTM-Z \cite{ghazi2019training} & 0.842$\pm$0.035$^*$ & 0.706$\pm$0.092 & 0.636$\pm$0.093 \\ 
    MinimalRNN \cite{nguyen2020predicting} & 0.871$\pm$0.015$^*$ & \textbf{0.743$\pm$0.091} & 0.644$\pm$0.083 \\
    DeepRNN \cite{jung2021deep} & 0.878$\pm$0.022$^*$ & {0.723$\pm$0.071} & 0.710$\pm$0.071 \\\midrule
    SPDSRU \cite{chakraborty2018statistical} & 0.776$\pm$0.049$^*$ & 0.655$\pm$0.0.024 & 0.563$\pm$0.093\\ 
    ManifoldDCNN \cite{zhen2019dilated} & 0.812$\pm$0.052$^*$ & 0.719$\pm$0.053 & 0.559$\pm$0.111 \\ \midrule
    Ours & \textbf{0.881$\pm$0.022} & 0.740$\pm$0.033 & \textbf{0.714$\pm$0.027}\\
    \bottomrule 
    \end{tabular}
    }
\end{table}

\begin{table*}[!t]
    \centering
    \caption{Performance of predicting cognitive scores in terms of MAPE and $R^2$. ($*$: $p<0.05$)}
    \label{exp:t2}
    \begin{tabular}{ccccccc}
    \toprule
    \multirow{2}{*}{Method} & \multicolumn{2}{c}{MMSE} & \multicolumn{2}{c}{ADAS-cog11} & \multicolumn{2}{c}{ADAS-cog13} \\ \cmidrule{2-7}
    & MAPE $\downarrow$ &$R^2$ $\uparrow$ & MAPE $\downarrow$ &$R^2$ $\uparrow$ & MAPE $\downarrow$ & $R^2$ $\uparrow$ \\
    \midrule\midrule
    LSTM-M & 0.173$\pm$0.030$^*$ & -0.412$\pm$1.143$^*$ & 0.929$\pm$0.433$^*$ & 0.321$\pm$0.173$^*$ & 0.863$\pm$0.289$^*$ & 0.302$\pm$0.267$^*$\\
    LSTM-F & 0.235$\pm$0.110$^*$ & -0.053$\pm$0.495$^*$& 0.829$\pm$0.353$^*$ & 0.198$\pm$0.494$^*$& 0.790$\pm$0.152$^*$ & 0.177$\pm$0.468$^*$ \\
    MRNN \cite{yoon2018estimating} & 0.149$\pm$0.031$^*$ & 0.168$\pm$0.284$^*$ &0.930$\pm$0.224$^*$ & 0.262$\pm$0.184$^*$& 0.920$\pm$0.234$^*$ & 0.263$\pm$0.187$^*$\\
    PLSTM-Z \cite{ghazi2019training} & 0.113$\pm$0.011$^*$ & 0.499$\pm$0.191& 0.575$\pm$0.121& 0.668$\pm$0.074& 0.566$\pm$0.136& 0.706$\pm$0.074\\
    MinimalRNN \cite{nguyen2020predicting} & 0.175$\pm$0.052$^*$ & 0.472$\pm$0.116$^*$ &0.565$\pm$0.142$^*$ & 0.569$\pm$0.038$^*$& 0.451$\pm$0.111& 0.635$\pm$0.049$^*$\\
    DeepRNN \cite{jung2021deep} & \textbf{0.082$\pm$0.012}& \textbf{0.683$\pm$0.102}&\underline{0.446$\pm$0.073}& \textbf{0.749$\pm$0.054}&\underline{0.422$\pm$0.092}& \textbf{0.777$\pm$0.050}\\
    Ours & \underline{0.099$\pm$0.018} & \underline{0.608$\pm$0.067}& \textbf{0.441$\pm$0.043} & \underline{0.689$\pm$0.036}& \textbf{0.403$\pm$0.043} & \underline{0.726$\pm$0.034}\\
    \bottomrule
    \end{tabular}
\end{table*}

\begin{table*}[!t]
    \centering
    \caption{Ablation study results (Manifold, Continuous).}
    \label{ab:t3}
    \begin{tabular}{c|ccc|cccc}
    \toprule
    Case & Method & Manfold & Continuous & mAUC & Recall & Precision \\
    \midrule\midrule
    % RNN-Decay
    Case I & ODE-RNN & & $\checkmark$ & 0.868$\pm$0.025 & 0.697$\pm$0.075 & 0.700$\pm$0.073 \\
    Case II & ODE-RNN + Dec. & & $\checkmark$ & \textbf{0.881$\pm$0.020} & 0.727$\pm$0.063 & \textbf{0.721$\pm$0.055} \\
    Case III & Ours (w/o ODE, Dec.) & $\checkmark$ & & 0.871$\pm$0.020 & 0.725$\pm$0.067 & 0.683$\pm$0.067 \\
    Case IV & Ours (w/o ODE) & $\checkmark$ & & 0.872$\pm$0.024 & 0.707$\pm$0.058 & 0.706$\pm$0.061 \\
    Case V & Ours (w/o Dec.) & $\checkmark$ & $\checkmark$ & 0.877$\pm$0.024 & 0.729$\pm$0.059 & 0.693$\pm$0.063 \\
    \midrule
     & Ours & $\checkmark$ & $\checkmark$ & \textbf{0.881$\pm$0.022} & \textbf{0.740$\pm$0.033} & 0.714$\pm$0.027\\
    \bottomrule
    \end{tabular}
\end{table*}

\subsection{Experimental Settings}
\subsubsection{RNN-based imputation methods}
A standard LSTM network with mean (LSTM-M) and forward (LSTM-F) imputations was employed for classification and regression tasks, such as independently predicting MRI biomarkers and cognitive scores. PLSTM-Z uses a peephole LSTM~\cite{gers2000recurrent} to impute missing values with zeros as input~\cite{ghazi2019training}. Therefore, we implemented PLSTM-Z to predict MRI biomarkers and cognitive scores and utilized an LDA classifier for classification. We also employed an MRNN that operates in inter- and intra-stream directions~\cite{yoon2018estimating}.
The MRNN’s output was applied as input for the LDA classifier’s task. Imputation modules directed MinimalRNN~\cite{chen2017minimalrnn} to extrapolate input features to impute missing values~\cite{nguyen2020predicting}. MinimalRNN simultaneously completed cognitive score prediction and classification. Next, a DeepRNN~\cite{jung2021deep} estimated missing values with an integrated LSTM network and imputation module that considers temporal and multivariate relations from input features. 

We initiated a hyperparameter search for RNN-based imputation methods using the following settings: hidden unit sizes, number of hidden layers, learning rate, and $\ell_2$-regularization $\{16, 32, 48, 64, 80, 96\}$, $\{1,2,3\}$, $5\times\{10^{-5},10^{-4},10^{-3},10^{-2}\}$, and $\{10^{-6},10^{-5},10^{-4},10^{-3}\}$, respectively. Early stopping was conducted to identify optimal hyperparameters for achieving the highest multi-class area under the receiver operating characteristic curve (mAUC) on the validation set. Lastly, we trained LSTM-M and LSTM-F with 64 hidden units and an Adam optimizer~\cite{kingma2014adam} with a $5\times 10^{-2}$ learning rate. For DeepRNN, MinimalRNN, MRNN, and PLSTM-Z, we set $5\times 10^{-3}$ learning rate and $\ell_2$-regularization with a coefficient of $10^{-4}$. We used a mini-batch size of 64, a single hidden layer, and 300 epochs for all models.

\subsubsection{Geometric learning-based methods}
ManifoldDCNN and SPDSRU models were implemented with an encoder $f_\theta$ for the covariance matrices and a shrinkage estimator~\cite{chen2010shrinkage} for a fair comparison. We then set the ManifoldDCNN and SPDSRU model output channels to 16 and 8, resulting in $16\times16$ and $8\times8$ covariance matrix dimensions, respectively.
We adhered to model settings established in a previous study~\cite{zhen2019dilated}. An Adam optimizer trained the ManifoldDCNN and SPDSRU models with $10^{-3}$ and $5\times 10^{-3}$ learning rates, respectively, and $\ell_2$ regularization with a weight coefficient of $10^{-4}$. We set a 64 mini-batch size and 300 epochs, respectively. Since neither method is designed to tackle missing values, only the prediction loss $\mathcal{L}_{\text{pred}}$ was considered.

\subsubsection{Proposed methods}  
Our proposed method incorporates two convolutional layers with a kernel size of 1, utilizing Batch normalization~\cite{ioffe2015batch} and LeakyReLU activation~\cite{maas2013rectifier}. The convolutional layer output channels were set to 32, resulting in a $32 \times 32$ SPD matrix dimension. Cholesky decomposition was applied to obtain a 32-dimensional vector for the diagonal component and a strictly lower $32\times (32-1)/2$-dimensional triangular component.
A 32-hidden unit size was selected for RGRU based on the hyperparameter search space. Using an Adam optimizer, we set a mini-batch size, epochs, and learning rate of 64, 300, and $10^{-3}$.
In addition, we applied $\ell_2$ regularization with a weight of $10^{-4}$ to prevent overfitting and achieve training loss convergence.

The hyperparameter search space for $\lambda_1$, $\lambda_2$, and $\lambda_3$ in the composite loss function \eqref{eq:18} was defined as $\{0.001, 0.01, 0.1, 0.2, \cdots, 1.0\}$, whereas the search space for $\delta$ in \eqref{eq:17} was set to $\{0,1,2,3,4,5\}$. We set $\lambda_1=1.0$, $\lambda_2=0.5$, and $\lambda_3=0.001$ values for \eqref{eq:18} and $\delta=5$ for \eqref{eq:17}.
Next, we determined the hyperparameters for the competing methods’ composite objective function in \eqref{eq:18}, which included the coefficient $\beta$ for imputation loss term and values of ($\beta=1.0, \lambda_1=0.1$) for LSTM-M and LSTM-F, ($\beta=0.25, \lambda_1=0.5$) for MRNN, ($\beta=1.0, \lambda_1=0.25$) for PLSTM-Z, ($\beta=1.0, \lambda_1=1.0, \lambda_2=1.0$) for MinimalRNN, and ($\beta=0.1, \lambda_1=0.5, \lambda_2=0.5$) for DeepRNN.

\subsection{Longitudinal Clinical Status Prediction}
\label{exp:sub_exp_result}
The validity of the proposed framework was demonstrated by evaluating its performance in a downstream task such as CN-versus-MCI-versus-AD classification over time for a maximum of ten-time points. We used the mAUC metric and five-fold cross-validation setting to evaluate our framework’s prediction task performance. As shown in TABLE~\ref{exp:t1}, our proposed method achieved significantly better mAUC and precision performance than the competing methods with $p<0.05$. The only recall exception was regarding MinimalRNN, which expressed a slightly better performance by 0.003. We observed the recall of our proposed framework was slightly lower is the high imbalance in sample sizes. Nevertheless, our proposed framework achieved a balanced recall and precision performance compared with other methods. For instance, our framework’s performance between these metrics was notably more balanced (a small gap; 0.026) than in MinimalRNN (0.099).

\begin{figure*}[!t]
\centering
    \includegraphics[width=\textwidth]{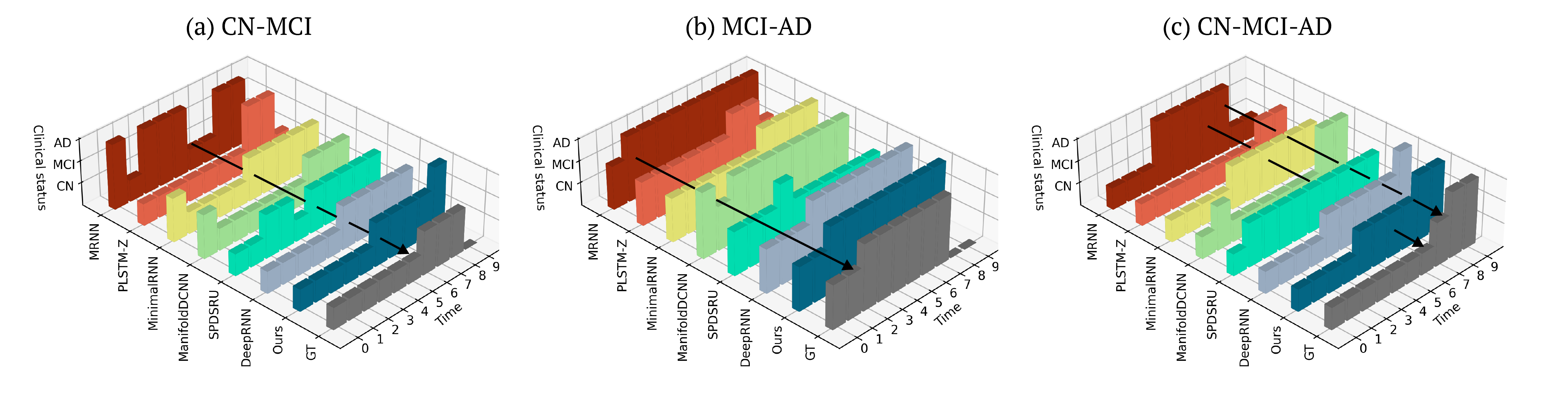}
    \caption{
    Result of a longitudinal status prediction comparison between the competing methods and our proposed method. Each row shows a transition of disease status from different subjects. The black arrow in each row indicates when a transition of disease status occurs. Note that GT denotes the ground truth.
    }
    \label{a1:transition}
\end{figure*}

\subsection{Cognitive Scores Prediction}
We also demonstrated the validity of the proposed framework by estimating the cognitive score prediction for the entire time sequence. Likewise, we carried out a five-fold cross-validation with the metrics of mean absolute percentage error (MAPE) and coefficient of determination ($R^2$). Furthermore, we implemented a statistical significance test between our framework and other comparative methods using the Wilcoxon signed-rank test~\cite{willcoxon1992individual}. Our proposed method obtained the highest MAPE for ADAS-cog11 and ADAS-cog13, as shown in TABLE~\ref{exp:t2}. Although our framework performed somewhat lower in several circumstances, no statistically significant differences were observed with p-values of 0.6 or higher, denoted by underlined results. 

\section{Analysis}
\label{sec:analysis}
We analyze our proposed framework in this section. We performed an ablation study to demonstrate the significance of each module. Further, we visualized predictive MRI biomarkers and longitudinal status predictions to observe how well our proposed method effectively captures the characteristics of irreversible neurodegeneration and sparsity in AD. Lastly, we demonstrated that the proposed method performs well in various settings and can effectively predict them.

\subsection{Ablation Study}
The efficiency analysis for each component in our proposed framework considered three essential aspects: temporal variability, sparsity, and geometrical properties (TABLE~\ref{ab:t3}). We compared our method with ODE-RNN~\cite{rubanova2019latent}, which implements an RNN to update hidden states in Euclidean space (Case I). Alternatively, our method utilizes RGRU and manifold space modeling to improve performance. Geometric characteristic impact in AD progression modeling was highlighted through this comparison, as manifold space modeling considerably enhanced our proposed method’s performance. We also discovered that incorporating a decoder to estimate missing values (Case II \& V) improved our model’s and ODE-RNN’s implementation. Further analysis revealed that the decoder influenced performance strikingly more than the ODE (Case V), corroborated by results from adding the ODE and decoder to Case III. We also noted that the ODE-based temporal modeling approach outperformed the discrete-time modeling (Case IV). Based on these findings, we concluded that our proposed model, which combines all three AD progression modeling aspects, outperforms existing models.

\subsection{Clinical Status Irreversibility}
\label{sec:irreversibility}
We compared our framework’s longitudinal status predictions with competing methods to evaluate its capacity for capturing clinical status irreversibility (Fig.~\ref{a1:transition}). Three different subjects with disease status transition were analyzed for comparison; MRNN, PLSTM-Z, ManifoldRCNN, and SPDSRU predictions exhibited substantial state-reversing errors, whereas MinimalRNN occasionally predicted state-reversals (Fig.~\ref{a1:transition}a). On the other hand, DeepRNN and our framework did not report any state-reversing errors. Despite MinimalRNN’s improved predictions (Fig.~\ref{a1:transition}b) compared to the other methods excluding DeepRNN, there were notable misclassifications (Fig.~\ref{a1:transition} (a,c)) and clinical status reversion errors (Fig.~\ref{a1:transition}a). Compared to DeepRNN, our proposed framework detected the disease status change earlier in two separate MCI and AD patients (Fig.~\ref{a1:transition} (b,c)).

\begin{figure}[!t]
    \centering
    \includegraphics[width=.5\textwidth]{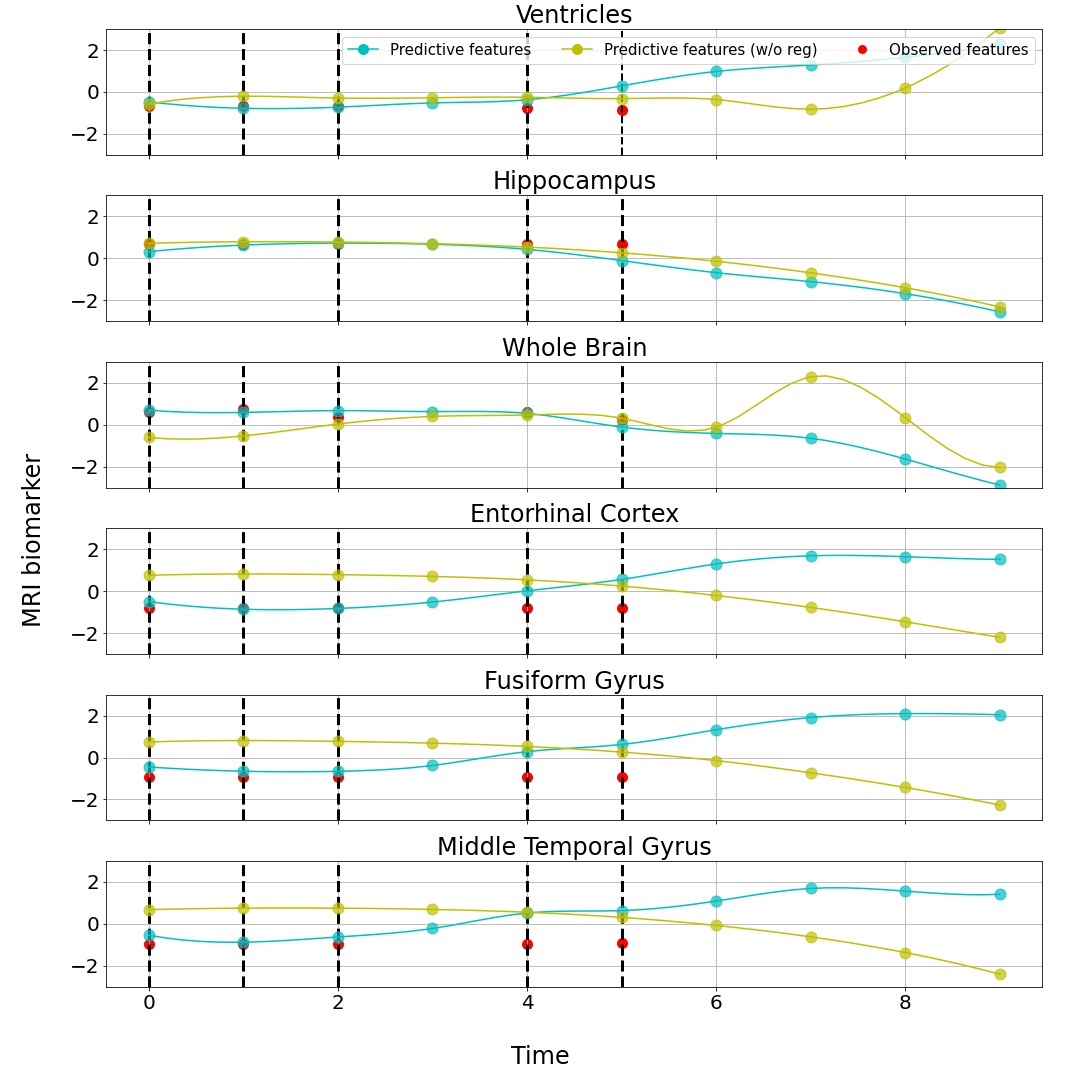}
    \caption{
    Predictive trajectories of MRI biomarkers over time. Each colored circle represents the predicted observations without/with regularization as a monotonicity constraint and the ground truth, and the resulting trend is shown as a solid line.}
    \label{fig:mri_biomarker}
\end{figure}

\begin{figure*}[!t]
\centering
    \includegraphics[width=\textwidth]{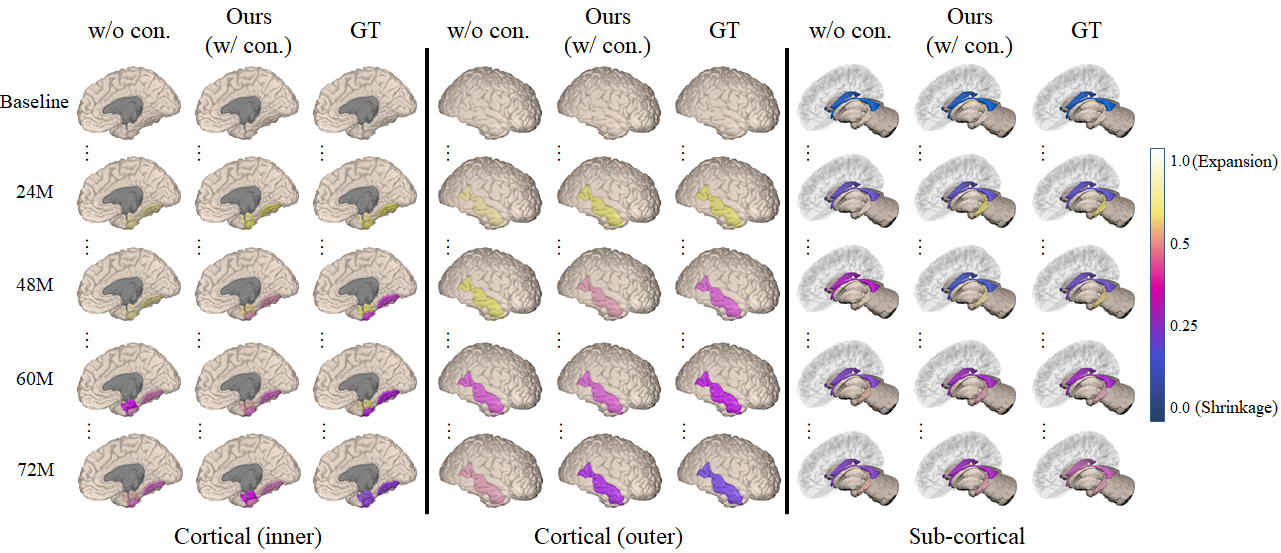}
    \caption{Visualization of 3D brain images depicting the predictive MRI biomarkers' trajectories over time. The visualizations include three scenarios: without monotonic constraint (w/o con.), with monotonic constraint (w/ con., ours), and the ground truth (GT).}
    \label{a1:transition2}
\end{figure*}

\subsection{Predictive MRI Biomarkers over Time}
We also evaluated our framework’s accuracy in predicting missing values with incomplete observations. Six MRI biomarker trajectories were predicted over time: entorhinal cortex, hippocampus, fusiform gyrus, middle temporal gyrus, ventricles, and whole brain (Fig.~\ref{fig:mri_biomarker}). Our proposed method exhibited precise missing observation predictions at time points 3, 6, 7, 8, and 9, where no observations were made across variables, and at 2 and 5, where incomplete observations were available. These results indicate that our proposed framework offers reliable predictions even with missing observations. 

Furthermore, we implemented a comparative analysis assessing our proposed method, which enforces monotonicity during the learning process, and models that do or do not consider regularization for prediction features. Although the optimization process without regularization still predicted observed features, some trajectories (ventricles and whole brain) presented results that were not medically feasible. In contrast, our results confirmed that optimization utilizing regularization improves monotonicity and accurately reflects AD progression’s irreversible nature, a critical aspect for precise AD progression modeling.

{\ws In addition, to intuitively understand, we visualized the 3D brain image representing predictive trajectories of MRI biomarkers under two distinct scenarios: one considering the monotonic constraint and the other without such consideration. We also included the visualization of the ground truth (GT) observations to ensure a fair comparison between the approaches (Fig.~\ref{a1:transition2}). Initially, we normalized the [0,1] range using the ground truth (GT) values from the baseline. Subsequently, the MRI biomarkers were mapped into specific brain regions for visualization. Specifically, the cortical (inner) regions included the entorhinal cortex and fusiform gyrus, the cortical (outer) region included the middle temporal gyrus, and the sub-cortical regions comprised the ventricle and hippocampus. The BrainPainter~\cite{marinescu2019brainpainter} was employed to make this figure, showcasing the predictive patterns of the MRI biomarkers in the designated brain areas.

% In Fig.~\ref{a1:transition2}, we observed that the scenario where we considered the monotonic constraint (ours) demonstrated predictions that closely resemble the ground truth (GT) when compared to the scenario without the constraint (w/o con.). 
{\sw In Fig.~\ref{a1:transition2}, we conducted a comparative analysis between two scenarios: one with the inclusion of the monotonic constraint (ours) and the other without the constraint (w/o con.). Our observations revealed that our proposed framework (w/ con.) resulted in predictions that closely resembled the ground truth (GT) when compared to the scenario without the constraint.}
However, we also noticed that our proposed framework (w/ con.) exhibits relatively accurate predictions up to 24 months, but from 48 months onwards, the prediction accuracy decreases compared to the previous time sequences. On the other hand, the scenario without the constraint (w/o con.) consistently {\sw showed} lower prediction accuracy compared to ours (w/ con.). Specifically, at 48 months, there was an excessive prediction in the sub-cortical area, and at 60 months, there was an overprediction in both the cortical (inner \& outer) areas. Furthermore, these overpredictions were also observed in the whole brain region, as depicted in Fig.~\ref{fig:mri_biomarker}. These observations indicate that considering the monotonic constraint results in more refined and accurate predictions. Despite some decline in accuracy after 24 months, our approach aligns better with the ground truth and offers more reliable predictions than the scenario without the constraint.
}

\begin{table}[!t]
    \centering
    \caption{Comparison of a multi-class classification task in different longitudinal scenarios (\eg, regular and irregular time settings).}
    \label{exp:t4}
    \scalebox{.95}{
    \begin{tabular}{cccc}
    \toprule
    Time interval & mAUC & Recall & Precision \\
    \midrule\midrule
    Irregular & \textbf{0.882$\pm$0.018} & \textbf{0.741$\pm$0.041} & \textbf{0.715$\pm$0.037}\\
    Regular & 0.881$\pm$0.022 & 0.740$\pm$0.033 & 0.714$\pm$0.027\\
    \bottomrule 
    \end{tabular}
    }
\end{table}

\subsection{Irregularly Sampled Time-Series Data}
Our proposed framework was also confirmed versatile by predicting clinical status in incongruous settings (\ie, irregular time). RNN-based approaches are applicable for modeling discrete-time dynamics with regular time intervals. Therefore, we incorporated the same settings from our baseline experiments by ignoring visiting months (e.g., 3, 6, and 18) and restricted the ADNI dataset to yearly follow-ups. Most real-world longitudinal data are irregular, and information loss occurs when regularly used with the conventional method. Comparatively, our proposed method manages every time point regardless of irregularity. Therefore, the entire TADPOLE dataset was evaluated to predict clinical status and biomarkers. Our proposed framework achieved slightly higher performance than other regular scenarios by incorporating more information (Table~\ref{exp:t4}).

\subsection{Multiple Time Point Predictions in Irregular Time Sequences}
\begin{figure*}[!t]
    \centering
    \includegraphics[width=.9\textwidth]{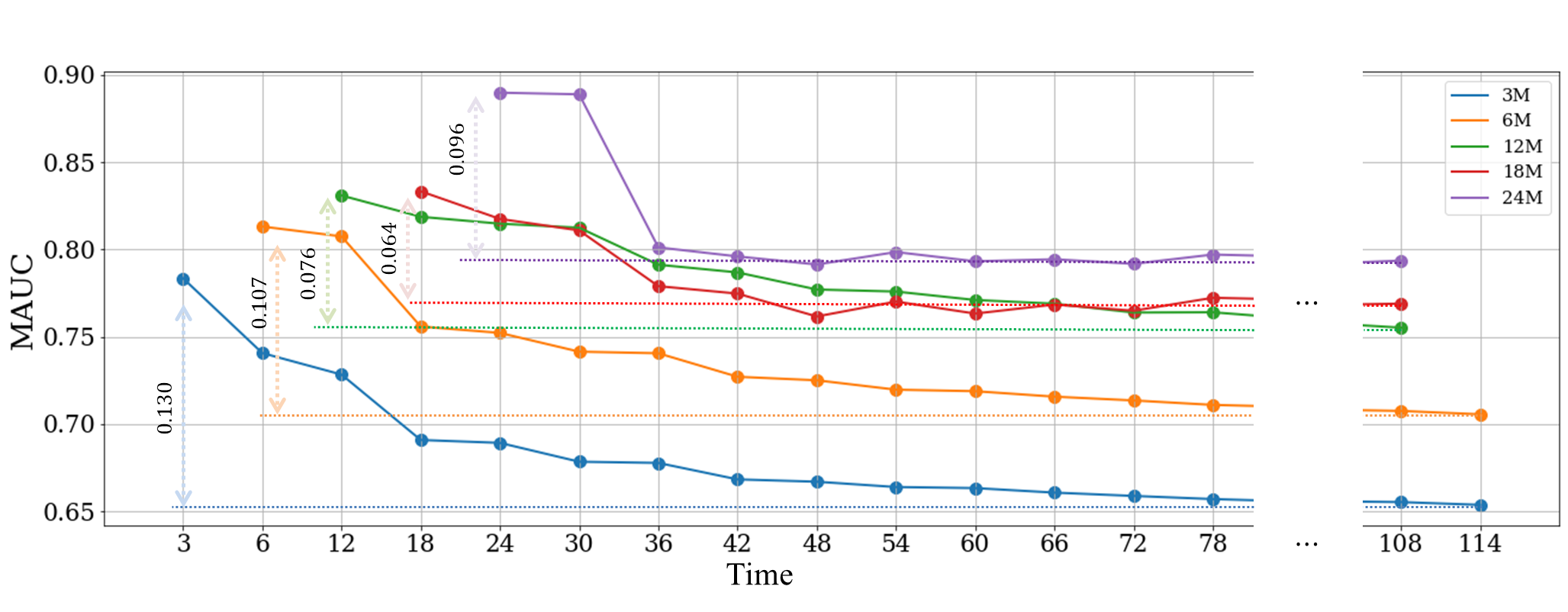}
    \caption{\ws
    Results of forecasting AD progression at multiple time points in an irregular setting. We utilized historical observations up to 24 months beyond the baseline (\eg, first visit) to forecast AD progression. The solid colored lines depict the disease prediction outcomes for the cohort using cumulative data up to that specific time point, with colors indicating the types of past observations used. In addition, we observe an improvement in both the first and last prediction results based on the degree of employing historical data.}
    \label{fig:forecasting}
\end{figure*}

We evaluated our proposed method regarding multiple time point predictions in an irregular time setting. Fig.~\ref{fig:forecasting} depicts the potential AD progression results beyond baseline predictions, {\ws utilizing irregularly accumulated data up to 24 months. For example, we utilized} two historical time points from baseline to three months {\ws to predict} clinical statuses over the relevant time points, {\ws while simultaneously extrapolating} MRI biomarkers and cognitive scores. The mAUC performance in predicting AD progression {\ws gradually improved as we incorporated additional historical data from each time point.
In this analysis, we observed a consistent and rapid decrease in data ratio beyond the 24 months. Consequently, we reported the potential prediction performance considering data only for up to 24 months. As a result, the performance difference based on the utilization of historical data was not significant from 36 months onwards. Notwithstanding this issue, we still observed an improvement in both the first and last prediction results as we employed more historical data. Specifically, when using historical data for up to 3 months, the difference in mAUC scores between the first and final prediction results was 0.130. For data up to 6 months, the difference was 0.107; up to 12 months, it was 0.076; up to 18 months, it was 0.064. Notably, when considering data up to 24 months, the difference increased significantly to 0.096 compared to the previous cases. Despite this notable increase, we observed that both the initial and final prediction results were improved compared to the former cases. This phenomenon was influenced by the rapid increase in the first prediction performance as we exploited more historical observations.  
Our findings are partially consistent with the patterns observed in previous studies using longitudinal data, emphasizing the importance of collecting and utilizing a substantial amount of historical observation data.}
% Furthermore, in Section~\ref{sec:irreversibility}, we validated the predictive accuracy of our proposed method concerning the conversion to AD for identifying the risk of AD progression within the specified time period.} 
% Furthermore, for early identification of AD progression risk, we also confirmed the predictive accuracy of our method regarding conversion to AD within the period of consideration in Section~\ref{sec:irreversibility}.

\section{Conclusion}
\label{sec:conclusion}
In this study, we devised a novel paradigm that harnesses geometric learning to model AD progression. The framework constitutes a topological space shift, ODE-RGRU, and trajectory estimation and successfully encapsulates temporal variability, observation sparsity, and geometric properties of temporal dynamics in regular and irregular settings. It also emulates the measurement irreversibility through monotonicity constraints during the optimization process. Our analysis revealed that each module in the proposed framework is integral for boosting performance. Some issues with estimating exact values exist, such as differentiating between estimation and true observation time points. Notwithstanding, our framework outperformed existing techniques in most parameters; however, further research is necessary to improve its clinical status prediction accuracy by better reflecting irreversible AD characteristics.

\section*{Acknowledgments}
This work was supported by Institute of Information \& communications Technology Planning \& Evaluation (IITP) grant funded by the Korea government(MSIT) (No. 2022-0-00959, (Part 2) Few-Shot Learning of Causal Inference in Vision and Language for Decision Making and No. 2019-0-00079 , Artificial Intelligence Graduate School Program (Korea University)) and by National Research Foundation of Korea (NRF) grant funded by the Korea government (MSIT) (No. 2022R1A2C2006865, Development of deep learning techniques for data-driven medical knowledge graph generation and interpretable multi-modal electronic health records analysis).

\bibliographystyle{IEEEtran}
\bibliography{main}

\end{document}